\documentclass[10pt,twocolumn,letterpaper]{article}
  
\usepackage{cvpr}              

\usepackage{graphicx}
\usepackage{amsmath}
\usepackage{amssymb}
\usepackage{booktabs}
\usepackage{xcolor}
\usepackage{textcomp}
\usepackage[pagebackref,breaklinks,colorlinks]{hyperref}

\usepackage[capitalize]{cleveref}
\crefname{section}{Sec.}{Secs.}
\Crefname{section}{Section}{Sections}
\Crefname{table}{Table}{Tables}
\crefname{table}{Tab.}{Tabs.}

\def\cvprPaperID{31} 
\def\confName{CVPR}
\def\confYear{2022}

\begin{document}

\title{Online Unsupervised Domain Adaptation for Person Re-identification}

 
\author{Hamza Rami$^{1,2}$,
        Matthieu Ospici$^{2}$,
        Stéphane Lathuilière$^{1}$\\
$^{1}$LTCI, Télécom Paris, Institut Polytechnique de Paris.\\
$^{2}$Atos.\\
\tt\small \{hamza.rami, stephane.lathuiliere\}@telecom-paris.fr, \tt\small matthieu.ospici@atos.net     
}

\maketitle
\begin{abstract}
  Unsupervised domain adaptation for person re-identification (Person Re-ID) is the task of transferring the learned knowledge on the labeled source domain to the unlabeled target domain. Most of the recent papers that address this problem adopt an offline training setting. More precisely, the training of the Re-ID model is done assuming that we have access to the complete training target domain data set. In this paper, we argue that the target domain generally consists of a stream of data in a practical real-world application, where data is continuously increasing from the different network's cameras. The Re-ID solutions are also constrained by confidentiality regulations stating that the collected data can be stored for only a limited period, hence the model can no longer get access to previously seen target images. Therefore, we present a new yet practical online setting for Unsupervised Domain Adaptation for person Re-ID with two main constraints: Online Adaptation and Privacy Protection. We then adapt and evaluate the state-of-the-art UDA algorithms on this new online setting using the well-known Market-1501, Duke, and MSMT17 benchmarks.
\end{abstract}



\section{Introduction}
\label{sec:intro}
Person Re-Identification (Re-ID) aims at recognizing a query of person-of-interest within a gallery of images. Therefore, we can determine whether this person has appeared in another place and/or has been captured by a different camera, or even the same camera at different times. The most typical scenario where Person Re-ID is used is video surveillance. A Person Re-ID model can be used to track people on a network of cameras or to find them given only a picture of them. Re-ID is therefore essential to many important surveillance applications.
\begin{figure}
  \centering
  \includegraphics[width=0.99\columnwidth]{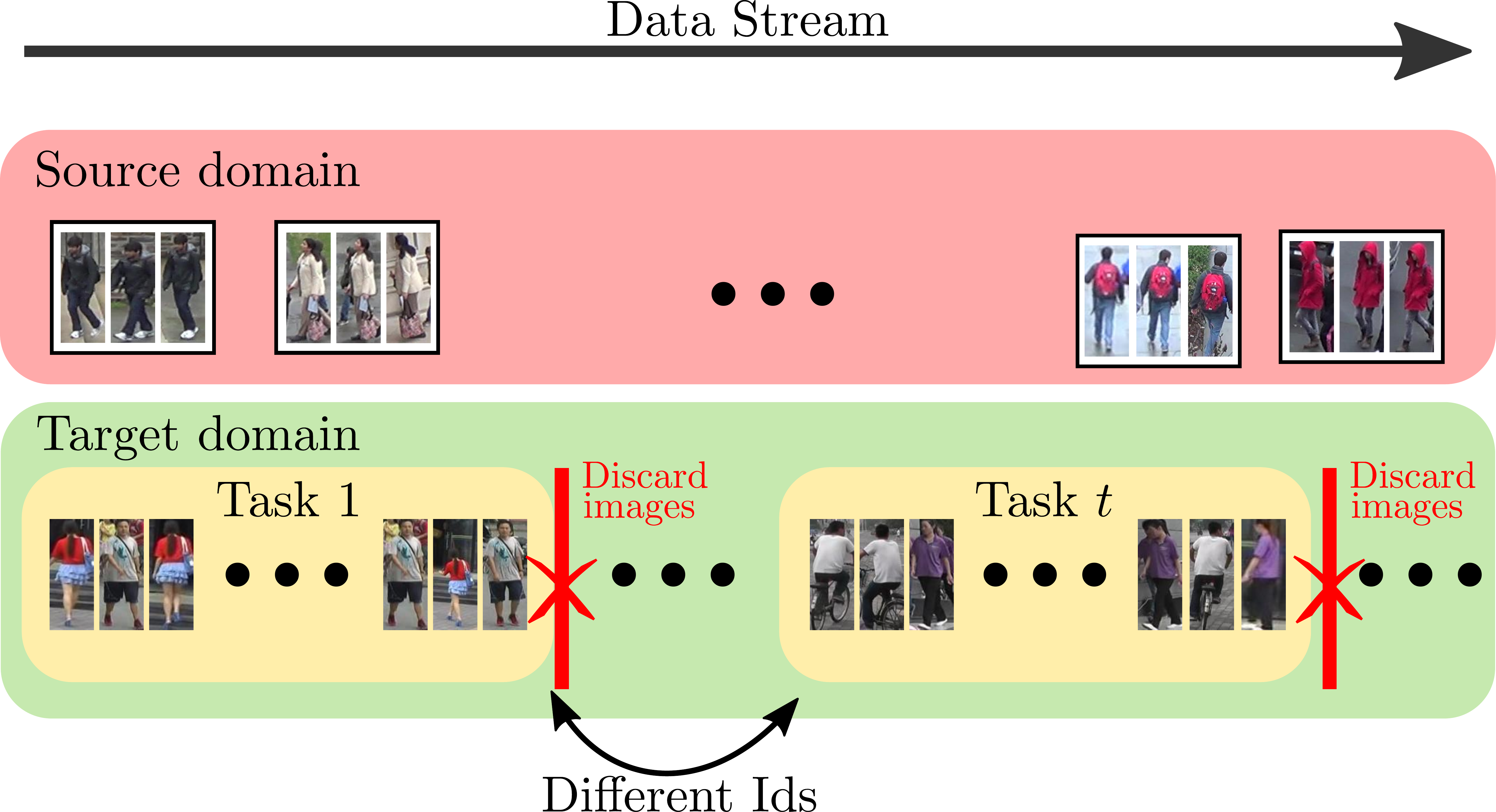}
    \caption{Illustration of the proposed OUDA for Re-ID setting: in Online Unsuperpervised Domain Adaptation for Person Re-ID, the annotated source dataset is available at any time while the target dataset is divided into annotated tasks. In between each task, the data from the previous task is discarded. }
      
  \label{fig:short}
\end{figure}

Recent advances in deep learning have improved the performance of the general Re-ID. Besides specialized architectures and algorithms, much of the recent progress can be attributed to the availability of large and annotated datasets\cite{Market,DukeMTMC,PTGAN}. As obtaining annotated data for Person Re-ID is tedious and costly, pre-trained models are often deployed in new test environments without any adaptation mechanism. However, these models, which work well when the test samples are collected in the same environment as the training data, suffer from important performance degradation when they are deployed in different locations or different camera networks.

To bridge this gap several \textit{Unsupervised Domain Adaptation} (UDA) approaches have been proposed in the literature. We can distinguish two major categories of UDA methods for Person Re-ID: domain translation-based methods \cite{PTGAN, SPGAN, HHL} and pseudo-label based methods \cite{SB, SSG, theory&practice, DGM, BottomUp, CPS}. Domain translation methods are based on the style transfer from the target domain to the source domain. Methods based on pseudo-labels generate pseudo-labels using clustering algorithms on the target domain and employ cluster index as class labels to perform adaptation \cite{SB, SSG, MMT,SpCL}.


Despite their relative efficiency, these UDA methods are all based on the assumption that we have access to a large set of samples from the target domain environment during the training to perform adaptation in an \emph{offline} fashion. In this work, we argue that this assumption is violated in many real-world scenarios. First, when deploying a Person Re-ID system, we gather images as long as they are recorded under the form of a stream that continuously provides images from different cameras/places. The \emph{offline} process implies that the Re-ID system requires a possibly long data collection phase before deployment. Second, since the Re-ID task evolves person identities, the system is confronted to confidentiality purposes in many countries, forcing the technology provider of such models to discard previously seen images. Therefore, we argue that to match practical scenarios, an unsupervised domain adaptation method for Person Re-ID should respect two main constraints: 1) \emph{Online adaptation:} the target domain data is not accessible all at once, but in a stream fashion where only small batches of images are available at a given instant of time and 2) \emph{Privacy protection}: Images captured by the different cameras can be used to update the Re-ID model and stored for only a limited period of time. To this end, in this paper, we propose and study a practical scenario for Unsupervised Person Re-ID which is the Online Unsupervised Domain Adaptation setting for Person Re-ID (OUDA-Rid). Fig. \ref{fig:short} gives an illustration of the proposed online setting, where we assume that the model has full access to the well-annotated source data set, however, unlike all the previous methods, the target domain dataset is fed to the Re-ID model in an online fashion. Practically, the target domain will be divided into several unlabeled subsets of images, where each subset will be viewed by the Re-ID model only once, hence respecting the two constraints: \emph{online adaptation} and \emph{privacy protection}. Regarding evaluation, we consider an independent and fixed target dataset with identities that do not overlap with any of the training tasks.

Our contributions can be summarized as twofold: 
\begin{itemize}
    \item We propose a new challenging yet practical scenario, the Online Unsupervised Domain Adaptation (OUDA) setting for Person Re-ID to respect two main constraints: Online Adaptation and Privacy Protection of identities.
    \item We adapt and evaluate three existing frameworks for \emph{offline} UDA to the proposed OUDA setting: the Strong Baseline \cite{SB}, MMT \cite{MMT} and SpCL \cite{SpCL}. These methods are evaluated in four different adaptation settings based on three public and widely-used datasets: Market 1501 \cite{Market}, DukeMTMC-reID \cite{DukeMTMC} and MSMT17 \cite{PTGAN}. Our results provide some interesting experimental conclusions regarding the performance and limitations of existing approaches. We hope that this work will stimulate the community to address domain adaptive Re-ID in the OUDA setting.
\end{itemize}


\section{Related Work}
\label{sec:related}

\noindent\textbf{Unsupervised domain adaptation (UDA) for person Re-identification} has been recently gaining a lot of attention for its practical applications. UDA methods can transfer learned knowledge from an annotated source domain to an unlabeled target domain, thus reducing the cost and discarding the need to have a well-annotated data set. Most of the existing methods and approaches in this area can be divided into two main categories: Domain translation-based and Pseudo label-based methods.

\textbf{Domain translation-based methods}  employ style transfer methods to modify the source images to obtain images with the content of the source domain but the appearance of the target. In this way, they obtain images similar to the target with the corresponding label annotations from the source images. These generated images are then used to refine the network parameters. Recent works in this category investigate the integration of generative models \cite{ATN} \cite{SDA} \cite{IGC}, as an example, we have \cite{PTGAN} which is based on CycleGAN \cite{CycleGan} to bridge the domain gap by transferring persons from the source domain to the target. \cite{SPGAN} also generates images while preserving the self-similarity of the images before and after the translation and the domain-dissimilarity of the translated source images to the target images. And finally, we can cite the work of Zhong et al. \cite{HHL} where the proposed framework learns camera-invariant features while enforcing domain connectedness, where two images, one from the source domain and the other one from the target domain, are fed to the network as a negative pair of images.

\textbf{Pseudo-labeling methods}, also called clustering-based methods, employ an iterative process alternating between clustering and finetuning \cite{theory&practice, DGM, BottomUp, CPS,delorme2021canu}. In its most simple implementation, \cite{SB}, the cluster indexes obtained in the cluster stage are used as labels to fine-tune the Re-ID network. Despite its simplicity, this simple approach obtains satisfactory results but suffers from limitations that have been addressed in recent works. For instance, Yang Fu et al. proposed a Self-Similarity Grouping (SSG) \cite{SSG} approach that assigns different pseudo-labels to both global and local features. To mitigate the effects of noisy hard pseudo-labels, Mutual-Mean Teaching (MMT) \cite{MMT}, proposed by Yaxiao et al., adopts a teacher-student framework with two networks that are trained jointly using hard pseudo labels generated by the two networks and soft pseudo labels generated by their Mean Networks, to conduct pseudo-label refinement in the target domain. Moreover, we can mention the work done by Ger et al referred to as SpCL \cite{SpCL} that, unlike previous methods, takes advantage of both labeled source domain images' centroids and un-clustered target instances, stored in a hybrid memory, in addition to the target domain clusters. The memory gives more supervision to the feature extractor during training while minimizing the unified contrastive loss over the three kinds of information available in the hybrid memory. 
Importantly, pseudo-label-based methods achieve better results than translation approaches and maintain up until now the state-of-the-art performances on almost all public datasets \cite{MMT,SpCL}. In addition, these approaches avoid the computation overhead of the transfer-based approach that requires image generation. Consequently, our experimental benchmark will focus only on pseudo-labeling approaches.

Even though all the aforementioned methods have shown promising results and great capability to adapt to a new target domain data set, their training process always assumes that they can have access to the entire target domain, which is difficult to hold in a real-world application as previously discussed in Sec.\ref{sec:intro}. 

\noindent\textbf{Lifelong Learning for person Re-Identification.} Lifelong Learning, also called Continual Learning or Incremental Learning\cite{Online1, Online2, Online3}, 
is a field that aims at mitigating the catastrophic forgetting problem, which means that the model tends to forget previous knowledge acquired during previous tasks when learning new ones.
Recently, many approaches have been developed to solve this problem for common vision tasks such as object detection \cite{onlineod}, segmentation \cite{onlineseg} or even image generation \cite{lifelonggan}. We can categorize existing methods into three main categories. First, teacher-student frameworks\cite{LWF, lifelongts}, use a teacher module to remind the student network about the knowledge acquired in the past. The second category of methods relies on the regularization of the parameters update when new tasks arrive \cite{lifelong_learning}. Finally, the third category is replay methods that consist in using stored images or an image generation model to feed old-task images along with the current task images into the learning network \cite{MRGAN}.

Recently, only a few works tackle the problem of lifelong learning in the case of Person Re-ID. \cite{AKA} propose an Adaptive Knowledge Accumulation (AKA) framework, however, the training process is fully supervised and treats only the domain-incremental scenario. Zhipeng Huang et al. \cite{lifelong_uda_reid} address a scenario similar to ours except that storing images from the previous task is permitted. In our work, we consider that in real-world applications, person images might be subject to confidentiality purposes, and therefore storing images from previous tasks is not permitted.

\section{Online Setting for UDA for Person Re-ID (OUDA-Rid)}
\label{sec:method}
\subsection{Problem Definition}
In this section, we describe the proposed online unsupervised domain adaptation setting for person re-identification (OUDA-Rid).
We consider that we have access to an annotated source domain data set  $D_S = \{(x_{i}^{S},y_{i}^{S})|_{i=1}^{Ns} \}$, where $x_{i}^{S}$ and $y_{i}^{S}$ denote the $i^{th}$ training sample and its associated person identity label. We consider that we also have access to a target domain data set $D_{T}$ where ground truth identity labels are not available. 
However, differently from the standard UDA setting, we consider that the target data set is accessible as an online stream of data. More precisely, we adopt the batch-based relaxation~\cite{fini2020online} of the online learning scenario. The model will have access to the target domain $D_{T}$ as a stream of $T$ independent batches $T_t,t\in{1..T}$. In analogy with the Continual Learning (CL) setting and to avoid confusion with the \emph{mini-batch} used in Stochastic Gradient Descent (SGD), each target batch will be called task. Each task $T_t$ is composed of $N_t$ images $\{x_{i}^{t}, i=1..N_t \}$ that depict an unknown number of identities. We assume that there is no identity overlap between tasks even if our approach does not strictly require it. This assumption corresponds to the practical scenario where data are collected over several hours or days. Even if the same person can appear again at different times, most detection will correspond to different identities. 

Importantly, we consider that at the end of the task, the images of the task $T_t$ cannot be used for the next tasks. This corresponds to a practical scenario where sensitive data can only be stored for a short period of privacy concerns (e.g. camera images from a public area). In addition to the source domain that is accessible at any time, only the parameters of the networks can be kept in memory in between two tasks. Finally, the goal is to deploy the trained model on an unknown target dataset that follows the same distribution as the training target tasks but does not share identities with the training tasks.

In this work, we adapt three frameworks for UDA to our OUDA setting. As detailed in Sec.~\ref{sec:related}, the UDA methods based on pseudo-labeling dominate most person re-identification benchmarks. Therefore, we focus our work on this paradigm. First, we employ a \emph{Strong baseline} that is a very simple, yet effective, baseline. Then, we consider MMT \cite{MMT} and SpCL \cite{SpCL}, which are two methods that achieve state-of-the-art performance on publicly available datasets. 
Apart from their performance, what motivates the choice of these two frameworks is that, on the one hand, MMT has attracted a lot of attention lately and it is now considered a reference baseline for the task of UDA for person re-identification. On the other hand, SpCL is  included in our benchmark since it illustrates the potential advantage of employing a memory to combine source and target data.
Once adapted, the three frameworks will be evaluated and tested under four different configurations to: 1) decide which of the three frameworks is most suited to the OUDA problem 2) measure the drop in performance due to the online constraint 3) study the sensitivity of each model to its hyper-parameters.

\subsection{Strong Baseline}
\begin{figure}
  \centering
  \includegraphics[width=0.8\columnwidth]{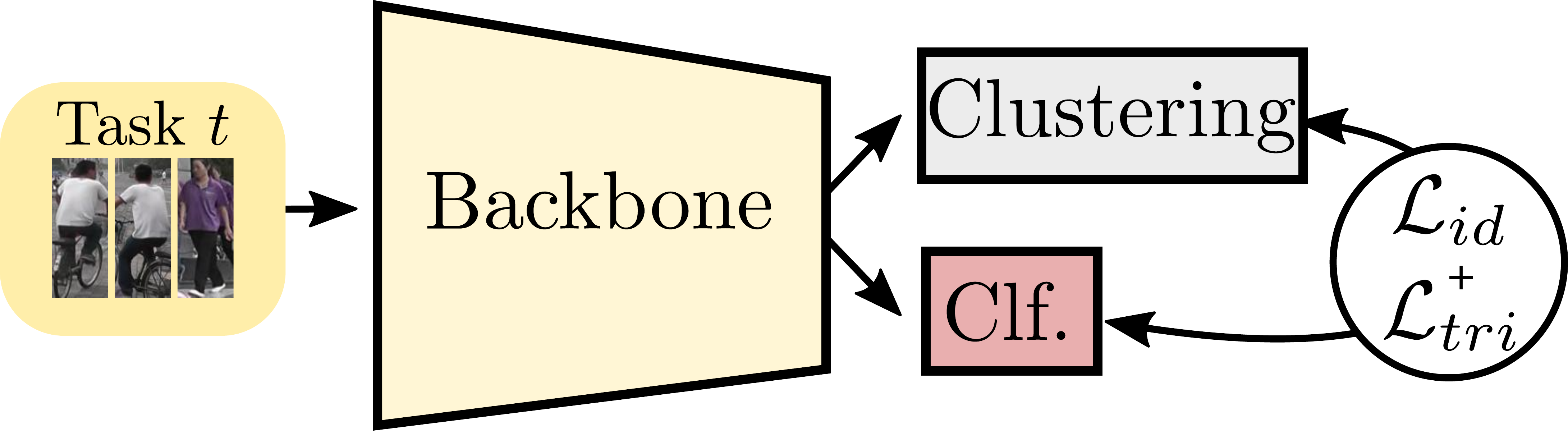}
    \vspace{-0.2cm}
    \caption{Scheme of \emph{Strong Baseline}: training iterate between clustering and finetunning. The network is trained using a combination of cross-entropy and triplet losses.}
  \label{fig:SB}
\end{figure}

The \emph{Strong Baseline} \cite{SB} is a simple pseudo labeling pipeline. A feature extractor network $F$ (backbone in Fig. \ref{fig:SB}) is pre-trained on the source labeled domain data set.  
After pretraining, the model is then fine-tuned on the target unlabeled data set.
The finetuning on target consists of an iterative process where two major steps are alternated until convergence:
\begin{enumerate}
    \item $F$ is used to extract image features for every target domain image. Then, a standard clustering algorithm (DBSCAN \cite{dbscan} in our experiments) is applied to the extracted target domain features to obtain $K$ clusters. In our case, $K$ is automatically returned by the DBSCAN algorithm. In this way, we assign a cluster label to every image.
    \item $F$ is then finetuned on the target samples using their cluster labels as pseudo-labels. More precisely, a target domain classifier $C$ with $K$ classes is added to classify the images' features along with their assigned pseudo labels. The network is then trained via the minimization of a combination of an identity loss $\mathcal{L}_{id}^{T}( \theta )$ and a triplet loss $\mathcal{L}_{tri}^{T}( \theta )$. Assuming a sample $x_i$ with pseudo-label $y_i$, the identity loss is given by:
      \begin{align}
        \mathcal{L}_{id} = \mathcal{L}_{ce}(C(F(x_{i})), y_{i}),
      \end{align}
      where $\mathcal{L}_{ce}$ denotes the cross-entropy loss. Assuming the hardest positive and hardest negative features in the current mini-batch for the sample $x_{i}$, denoted $f_i^{+}$ and $f_i^{-}$ respectively, the triplet can be written:      
\begin{align}
  \mathcal{L}_{tri}^{T}( \theta ) =  \max [ 0,&||F(x_i) -f_i^{+}) ||\notag\\
  &+ m - ||F(x_i) -f_{i}^{-}) || ]
  \end{align}
\end{enumerate}
where $||.||$ denotes the $\mathcal{L}^2$-norm and $m=0.5$ denotes the triplet distance margin.\\

\noindent\textbf{Adaptation to OUDA.} 

In our setting, this baseline approach is applied to each task. Instead of using the whole target dataset in the clustering step, we use only the data of the current task. The two steps are applied iteratively for several epochs.

\subsection{MMT}

\begin{figure}
  \centering
  \includegraphics[width=0.99\columnwidth]{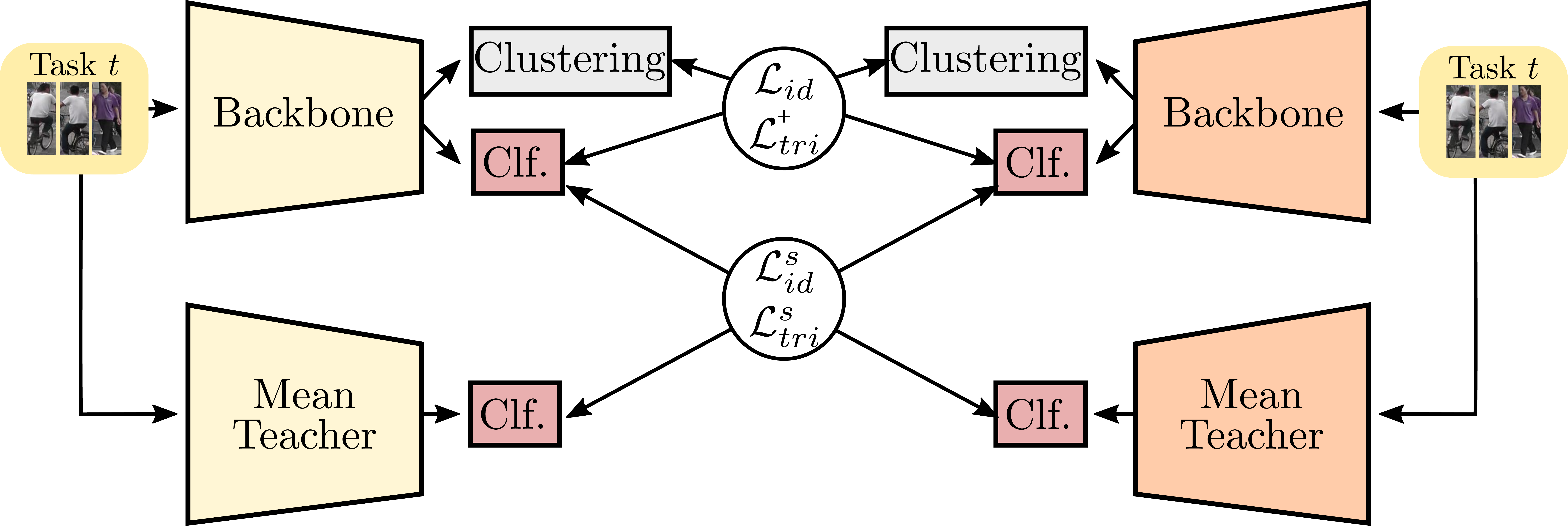}
      \vspace{-0.2cm}
    \caption{Scheme of \emph{MMT}: two networks are trained thanks to two other momentum encoder networks. The two networks are trained using a combination of cross-entropy and triplet losses.}
  \label{fig:MMT}
\end{figure}

MMT is a recent framework proposed by \cite{MMT}, that integrates the teacher-student framework with two networks that train jointly. The main motivation is to design a framework that uses both hard and soft pseudo labels to learn better features.
As shown in Fig.\ref{fig:MMT}, MMT extends the \emph{Strong Baseline} in several ways. First, MMT employs two networks $F_1$ and $F_2$ instead of a single feature extractor $F$. To enforce that the networks help each other, the classifier $C_1$ for the feature extractor $F_1$ is trained to predict the clustering labels obtained from $F_2$ and vice-versa. Second, mean teacher networks  $M_1$ and $M_2$ are introduced. These networks are obtained via estimating the running average on the network parameters of $F_1$ and $F_2$. These networks predict more stable pseudo labels since they combine the knowledge of the networks at previous training iterations.
In addition to the losses identity and triplet losses introduced in the \emph{Strong baseline}, the two networks $F_1$ and $F_2$ are also optimized with respect to a soft classification loss and a soft triplet loss. Those losses are calculated for each network over the predictions of the other mean network. The losses between $F_1$ and $M_2$ are:
\begin{align}
  \mathcal{L}_{sid} = -  M_2(x_i ). \log C_1(F_1(x_i))\\ 
  \mathcal{L}_{stri} = - \mathcal{L}_{bce} ( \tau_1^F (x_i), \tau_2^M (x_i)),
\end{align}
where $\mathcal{L}_{bce}$ denotes the binary cross entropy and $\tau_1^F$ and $\tau_2^M$ are given by: 
\begin{align}
\tau_1^F(x_i) = \frac{ e^{||F_1(x_{i})\!-\!F_1(x_{i}^{-}) ||}}{ e^{||F_1(x_{i})\!-\!F_1(x_{i}^{+}) ||} + e^{||F_1(x_{i})\!-\!F_1(x_{i}^{-} ) ||} }
\end{align}
\begin{align}
 \tau_2^M(x_i) = \frac{ e^{||M_2(x_{i})\!-\!M_2(x_{i}^{-}) ||}}{ e^{||M_2(x_{i})\!-\!M_2(x_{i}^{+}) ||} + e^{||M_2(x_{i})\!-\!M_2(x_{i}^{-} ) ||} }
\end{align}
Note that to encourage the two networks to learn different image representations, different random data transformation policies are used for each network pairs $(F_1, M_1)$ and $(F_2, M_2)$.\\

\noindent\textbf{Adaptation to OUDA.} 

We adapt MMT to the OUDA setting in the following way: at the end of each task, the parameters of the four networks are kept and reused for the next task.

\subsection{SpCL}
\begin{figure}
  \centering
  \includegraphics[width=0.99\columnwidth]{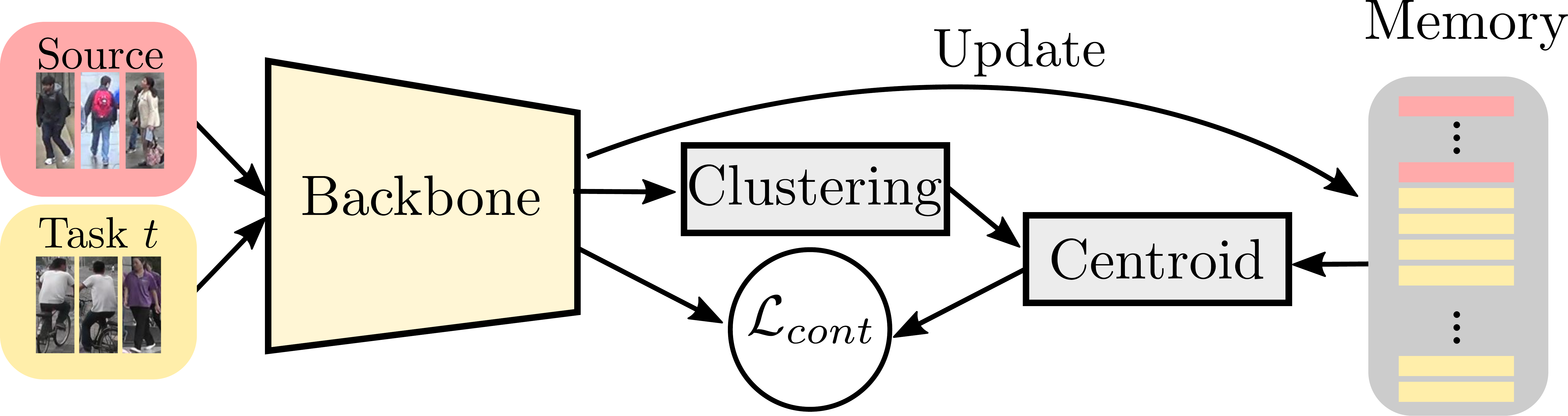}
      \vspace{-0.2cm}
    \caption{Scheme of \emph{SpCL}: a feature memory is used to perform contrastive learning.}
  \label{fig:spcl}
\end{figure}

Finally, we consider the SpCL method proposed in~\cite{SpCL}. This framework (Fig.~\ref{fig:spcl}) employs a hybrid memory that stores and continually updates three types of feature vectors: 
the class centroids for every class of the source-domain, cluster centroids for every cluster from the target domain, and the image feature corresponding to the target-domain samples that are not assigned to any cluster and that are considered as outliers. This memory provides supervision to the feature extractor via a contrastive loss over the three types of features in the memory.\\

\noindent\textbf{Adaptation to OUDA.}

We consider two adaptations of the SpCL framework. In the first version, referred to \emph{SpCL-SF}, we adopt a source-free SF strategy~\cite{kundu2020universal,saltori2020sf} where we do not use any of the source data when adapting to the target domain. This version is introduced because it allows for a fair comparison with MMT and the \emph{Strong Baseline} that use the source dataset only for pretraining. In our second version (simply referred to as \emph{SpCL}), we use the source dataset during the whole adaptation process in addition to the target data of the current task. In both cases, the memory is emptied, and clustering is performed at the beginning of each task.



\section{Experiments}
\label{sec:expes}
\subsection{Datasets}
We evaluate the different frameworks on three widely-used real-world person benchmark datasets in Domain Adaptive Person Re-ID:
\begin{itemize}
    \item Market 1501 \cite{Market}: is a large-scale public dataset that contains 1501 identities that are captured by six different cameras. The total number of images is 32,668 for which 12,936 images of 751 identities are used for training and 19,732 images corresponding to the remaining 750 identities are used as a test set. We follow the official testing protocol stating that 3,368 query images should be tested and matched to 19,732 gallery images.
    \item DukeMTMC-reID \cite{DukeMTMC}: The Duke Multi-Tracking Multi-Camera Re-Identification consists of images extracted from videos captured by 8 different cameras. It contains 16,522 training images corresponding to 702 identities, 2,228 query images of another 702 identities along 17,661 gallery images for testing.
    \item MSMT17 \cite{PTGAN}: The third benchmark is the most challenging dataset since it has a greater diversity in terms of people’s appearances, viewpoints, and scales. It consists of multiple hours of videos captured by 15 different cameras. This dataset is a large-scale dataset consisting of 32,621 images of 1,042 identities as a training set, and 11,659 query images along with 82,161 gallery images corresponding to 3,060 identities as a test set.
\end{itemize}

\begin{figure*}[t]
  \begin{minipage}[c]{0.33\textwidth}
    \centering
    \begin{tabular}{llcccccc}\toprule
{}                 & \multicolumn{2}{c}{\bf Offline} & \multicolumn{2}{c}{\bf Direct inference} & \multicolumn{2}{c}{\bf Online} \\ \midrule
{} & {mAP} & {Rank-1} & {mAP} & {Rank-1} & {mAP} & {Rank-1}\\ 
{Strong Baseline} & {75.6} & {90.9}   & {29.6} & {62.4}           & {49.4} & 77.1   \\ 
{MMT}             & {80.9} & {92.9}    & {29.6} & {62.4}           & {63.7} & {87.5}     \\
{{SpCL-SF}}            & {76.7} & {90.3}    & {29.6} & {62.4}      & {42.9} & {70.2}  \\ 
{{SpCL }}   & {78.2} & {90.5}    & {29.6} & {62.4}      & {47.9} & {72.9}     \\ 
\bottomrule
  \end{tabular}
  \end{minipage}
  \hfill
\begin{minipage}[c]{0.40\textwidth}
    \centering

  \includegraphics[width=0.99\textwidth]{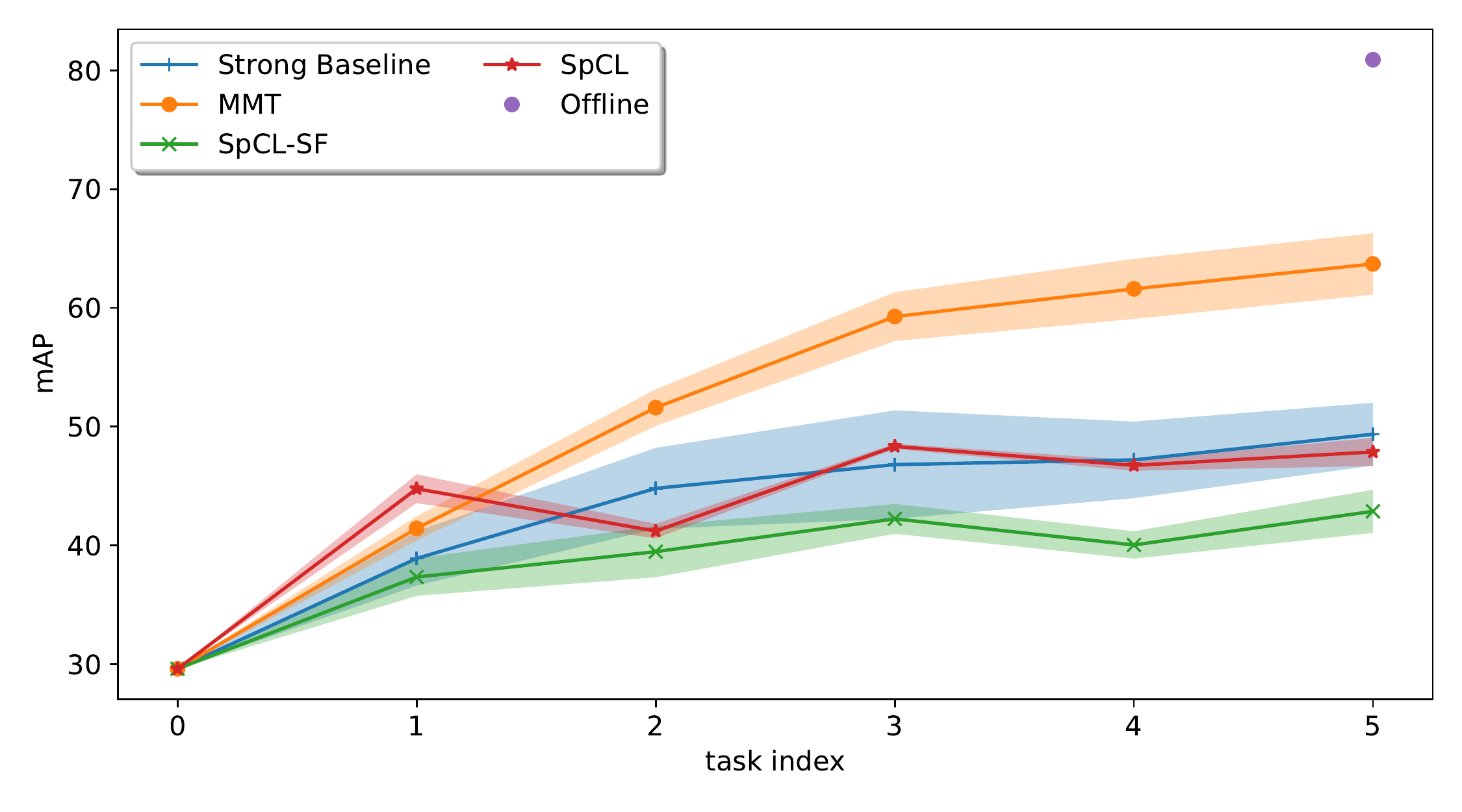}
\end{minipage}
  \vspace{-0.2cm}
\caption{Experimental comparison of the performance of the four methods (\emph{Strong baseline}, MMT, SpCL, and SpCL-SF) in the 5-task OUDA Duke -\textgreater Market configuration. We report mAP and Rank-1 accuracy for each method.}

\label{DM}

  \begin{minipage}[c]{0.55\textwidth}
    \centering
  \begin{tabular}{llcccccc}\toprule
{}                 & \multicolumn{2}{c}{\bf Offline} & \multicolumn{2}{c}{\bf Direct inference} & \multicolumn{2}{c}{\bf Online} \\ \midrule
{} & {mAP} & {Rank-1} & {mAP} & {Rank-1} & {mAP} & {Rank-1}\\
{{Strong Baseling}} & {60.4} & {75.9}    & {28.2} & {50.1}           & {26.8} & {59.3}   \\ 
{MMT}             & {67.7} & {80.3}   & {28.2} & {50.1}           & {51.7} & {72.3}  \\
{{SpCL-SF}}            & {68.8} & {82.9}    & {28.2} & {50.1}           & {38.7} & {61.8}  \\ 
{SpCL}   & {70.4} & {83.8}   & {28.2} & {50.1}         & {42.7} & {66.0}   \\ 
\bottomrule
  \end{tabular}
  \end{minipage}
  \hfill
  \begin{minipage}[c]{0.40\textwidth}
        \centering
\includegraphics[width=0.99\textwidth]{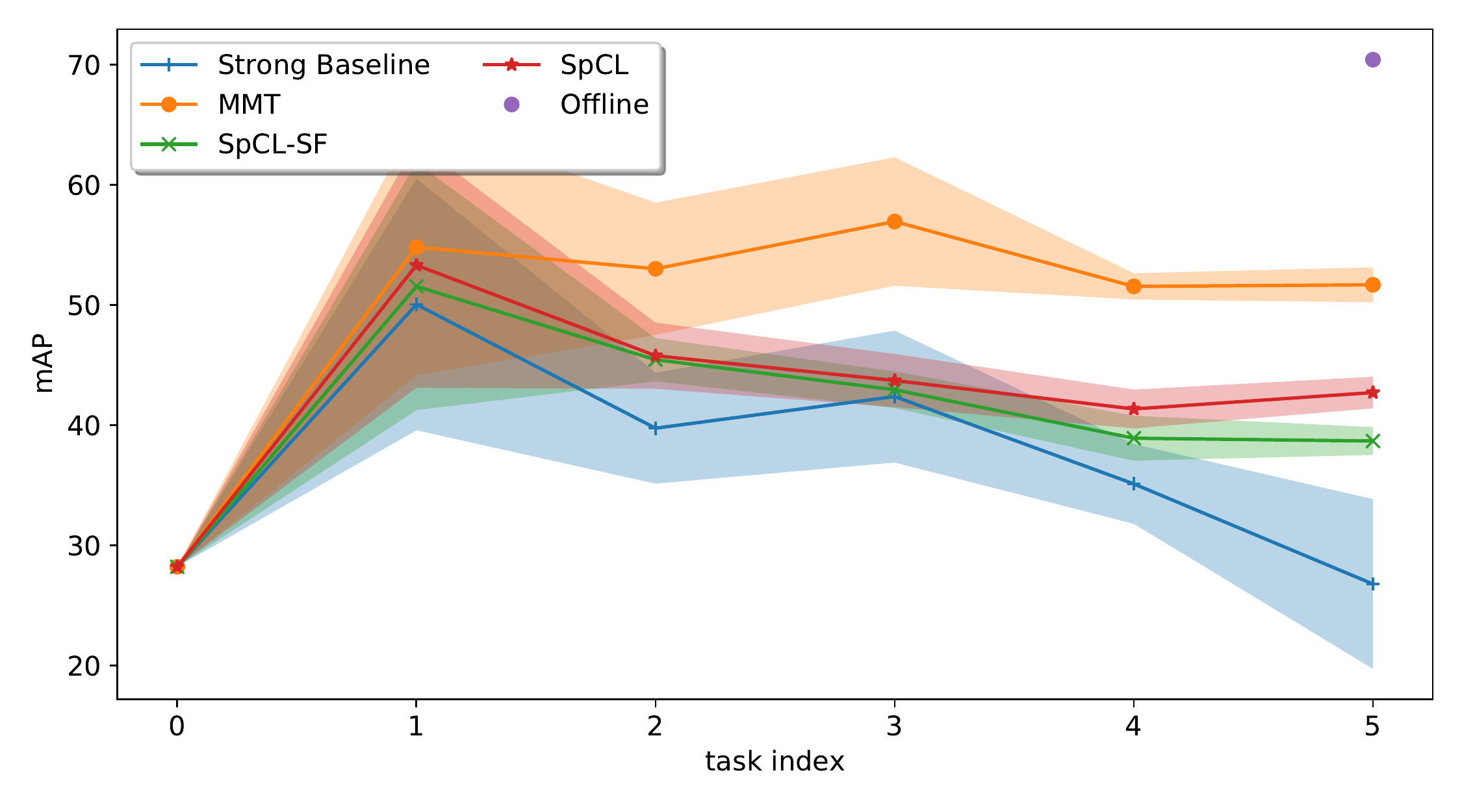}
\end{minipage}
    \vspace{-0.2cm}
  \caption{Experimental comparison of the performance of the four methods (\emph{Strong baseline}, MMT, SpCL, and SpCL-SF) in the 5-task OUDA Market -\textgreater Duke configuration. We report mAP and Rank-1 accuracy for each method.}
  \label{MD}

  \begin{minipage}[c]{0.55\textwidth}
    \centering
 \begin{tabular}{llcccccc}\toprule
{}                 & \multicolumn{2}{c}{\bf Offline} & \multicolumn{2}{c}{\bf Direct inference} & \multicolumn{2}{c}{\bf Online} \\ \midrule
{} & {mAP} & {Rank-1} & {mAP} & {Rank-1} & {mAP} & {Rank-1}\\
{{Strong Baseling}} & {9.7} & {25.8}      & {8.9} & {28.9}             & {6.1} & {18.0}   \\ 
{{MMT}}             & {22.9} & {49.2}  & {8.9} & {28.9}     & {15.1} & {31.5}   \\
{SpCL-SF}            & {26.3} & {53.4}   & {8.9} & {28.9}            & {13.1} & {36.5}  \\ 
{SpCL}   & {26.8} & {53.7}   & {8.9}  & {28.9}            &   {14.7} & {36.7}   \\ 
\bottomrule
\end{tabular}
  \end{minipage}
  \hfill
  \begin{minipage}[c]{0.40\textwidth}
        \centering
\includegraphics[width=0.99\textwidth]{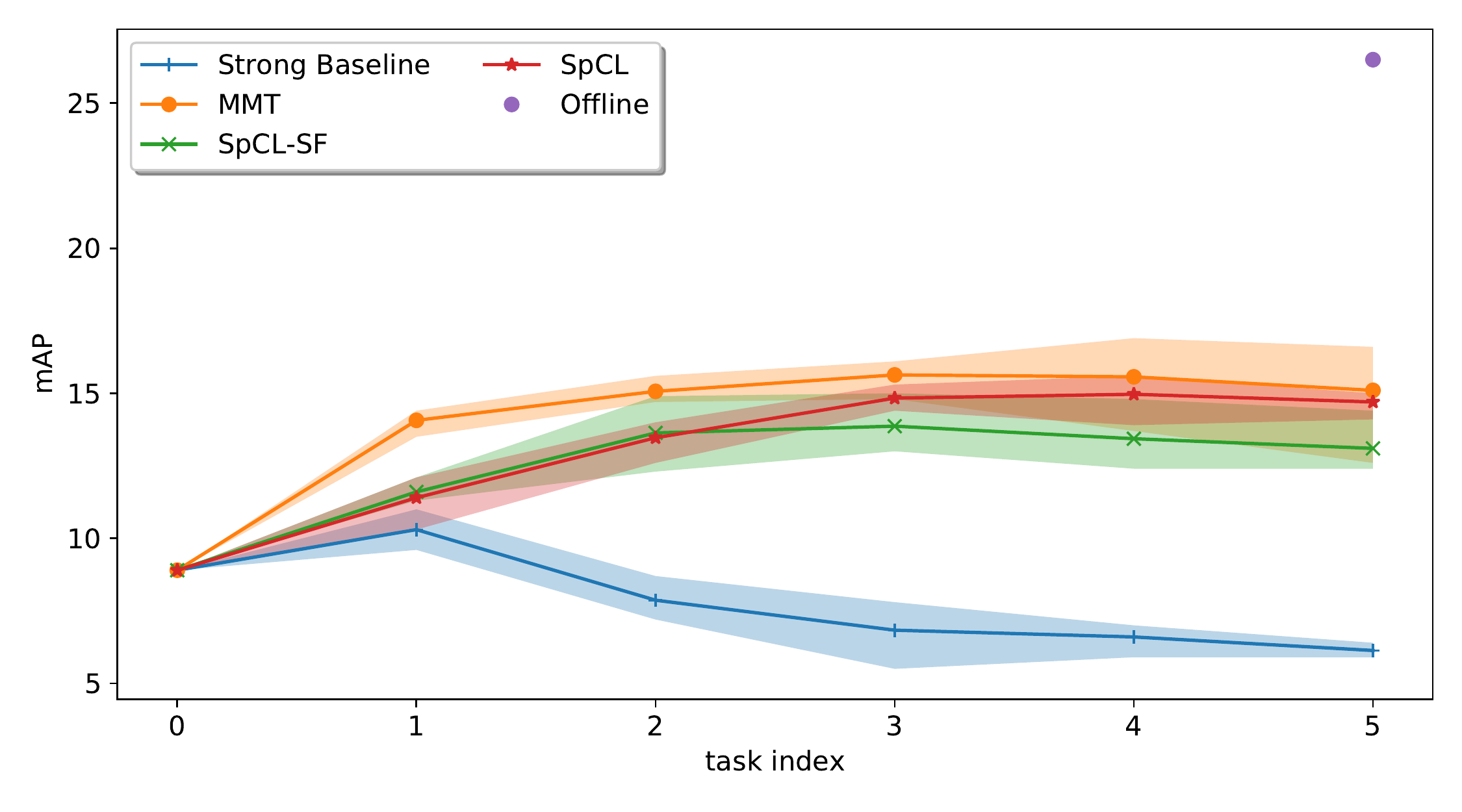}
\end{minipage}
  \vspace{-0.2cm}
\caption{Experimental comparison of the performance of the four methods (\emph{Strong baseline}, MMT, SpCL, and SpCL-SF) in the 5-task OUDA Market -\textgreater MSMT configuration. We report mAP and Rank-1 accuracy for each method.}
\label{fig:MMSMT}

  \begin{minipage}[c]{0.55\textwidth}
    \centering
\begin{tabular}{llcccccc}\toprule
{}                 & \multicolumn{2}{c}{\bf Offline} & \multicolumn{2}{c}{\bf Direct inference} & \multicolumn{2}{c}{\bf Online} \\ \midrule
{} & {mAP} & {Rank-1} & {mAP} & {Rank-1} & {mAP} & {Rank-1}\\
{{Strong Baseline}} & {10.9} & {28.6}       & {11.1} & {35.2}         & {7.2} & {19.9}    \\ 
{{MMT }}             & {23.3} & {50.1} & {11.1} & {35.2}     & {17.0} & {35.0}    \\
{SpCL-SF}            & {26.3} & {52.6}    & {11.1} & {35.2}         & {17.1} & {43.1}  \\ 
{SpCL}   & {26.5} & {53.1}   & {11.1} & {35.2}      & {17.8} & {40.8} \\ 

\bottomrule
\end{tabular}
  \end{minipage}
  \hfill
  \begin{minipage}[c]{0.40\textwidth}
        \centering

\includegraphics[width=0.99\textwidth]{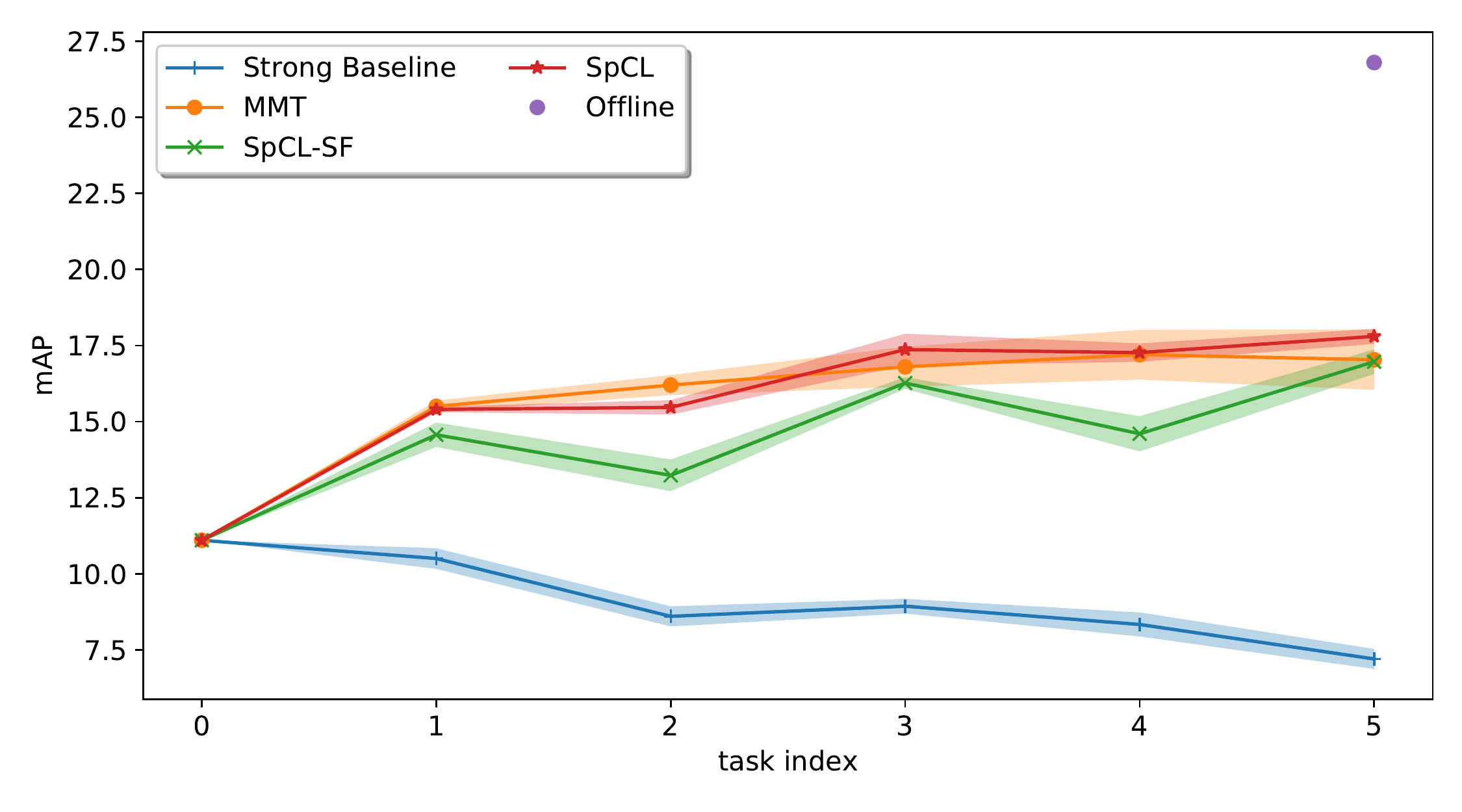}
\end{minipage}
  \vspace{-0.2cm}
\caption{Experimental comparison of the performance of the four methods (\emph{Strong baseline}, MMT, SpCL, and SpCL-SF) in the 5-task OUDA Duke -\textgreater MSMT configuration. We report mAP and Rank-1 accuracy for each method.}
\label{fig:DMSMT}
\end{figure*}

\subsection{Evaluation Protocol}
To evaluate the performance of the different frameworks on our proposed setting, we consider four source-target configurations:  Duke$\rightarrow$Market, Market$\rightarrow$Duke, Market$\rightarrow$MSMT17, and Duke$\rightarrow$MSMT17. These configurations are widely used in the literature and illustrate domain shifts of diverse difficulty. For each configuration, we randomly and uniformly split the training identities into 5 subsets corresponding to 5 tasks. For the evaluation metrics, we adopt the metrics commonly used in Re-ID \cite{MMT, SpCL}: Mean Average Precision (mAP) and Rank-1 \cite{cmc} accuracies. These metrics are computed on the entire test set of the target domain after each task during the online adaptation process. The proposed testing protocol is chosen to have a global overview of the model's adaptation capability to the domain shift between source and target, and also to see which framework is the most suited for online adaptation.

In our preliminary experiments, we observed that the number of epochs per task is a key hyper-parameter.
Even with a separate validation set, this hyper-parameter could not be chosen by mAP maximization since it requires identity labels and it would break the unsupervised adaptation assumption. On the contrary, using an inappropriate hyper-parameter value would jeopardize the validity of the conclusions of our experiments. Therefore, we used the following procedure: we run the \emph{strong baseline} with four different numbers of epochs ranging from 10 to 40. Then, we observed that training for 20 epochs per task leads to the best performance. Therefore, we use 20 epochs per task for all the methods. Note that we report an ablation study in Sec.\ref{sec:ablation} to measure the sensitivity to this hyper-parameter and we validate that this choice remains satisfactory for the other methods.

\subsection{Additional Baselines}

To better assess the performance of the evaluated approaches, we consider two additional baselines that are not trained following our OUDA setting. First, we report the performance of the model pre-trained on the source and directly evaluated on the target. This baseline is common to all the frameworks since all the methods use the same pre-trained model and it is referred to as \emph{Direct inference}. The second baseline is specific to each framework. It corresponds to the original method trained in the standard UDA setting and is referred to as \emph{Offline}. It can be interpreted as an upper bound for the online methods.

\subsection{Implementation details}
We follow the common practices in the UDA Person Re-ID field by adopting ResNet50 \cite{resnet} pre-trained on ImageNet \cite{imagenet} as a backbone. For clustering, we use DBSCAN \cite{dbscan} which is frequently used in the pseudo-label based methods as it requires no prior on the number of clusters but only the maximum distance between two samples to consider one in the neighborhood of the other. We employ the maximum distance hyper-parameter set in the original papers of MMT and SpCL. Adam \cite{adam} optimizer is adopted with an initialized learning rate equal to $3.5*10^{-4}$ and a weight decay of $0.0005$ \cite{MMT, SpCL}. Finally, all the images are resized to 256 x 128 before being fed into the backbone (backbones for MMT), and the batch size was set to 64 corresponding to 16 different identities with 4 images per ID.

\subsection{Results}
We report in Figs. \ref{DM}, \ref{MD}, \ref{fig:MMSMT} and \ref{fig:DMSMT} the results of the \emph{Strong Baseline} \cite{SB}, MMT \cite{MMT} and SpCL \cite{SpCL} on respectively four OUDA configurations: Duke$\rightarrow$Market, Market$\rightarrow$Duke, Market$\rightarrow$MSMT17 and Duke$\rightarrow$MSMT17. For every configuration, we report the final performances of each framework at the end of the adaptation process and plot the evolution of the test performance while the model is adapting to the target domain. Each experiment was repeated 3 times with different batch sampling initializations (\ie seeds). The colored area corresponds to the variance of the performance on the test set at the end of each task, where the points correspond to the mean performances of the different initializations.

First of all, in the four configurations, the results show that the pre-trained ResNet50 on the source domain gives poor performances when directly deploying it into the target domain without any finetuning (\emph{Direct inference}) compared to when it is fine-tuned on the target domain, either in an \emph{Offline} or \emph{Online} setting. This big gap in terms of performance illustrates the problem of domain shift.

Then, when it comes to the 5-tasks Online setting, the conclusions differ between methods and datasets. In the case of the Duke$\rightarrow$Market configuration (Fig.~\ref{DM}), we observe that MMT (orange line) performs best among the online methods and reaches $63.7\%$ of mAP. This result is very satisfactory since MMT bridges most of the gap between \emph{Direct inference} and \emph{Offline}. The \emph{strong baseline} obtains lower performance since it plateaus after the second task. However, it surprisingly outperforms the two SpCL variants. Indeed, the performance is not improved significantly after completing the first task. We even observe a small drop when processing the second task for the SpCL variant that uses the source domain images. We also notice that the difference between the two variants of SpCL is minor illustrating that with a straightforward adaptation of the SpCL method, initially proposed for offline UDA, SpCL does not benefit much from the availability of the source data. 
On the right-hand side of the Fig.\ref{DM}, we can see that the \emph{Strong Baseline} shows the highest sensibility to random seeds. Moreover, MMT keeps reaching the best performance independently of the random seed.

Regarding the Market$\rightarrow$Duke configuration (see Fig.~\ref{MD}), MMT is again the best performing method even though its gap with respect to the best offline method (purple dot) is larger.  
This behavior change can be explained by the highest difficulty of this setting as illustrated by the lower score obtained by the offline methods (\eg $70.4\%$ of the map in Market $\rightarrow$Duke vs $80.9\%$ of the map in Duke $\rightarrow$Market). In this more difficult configuration, the \emph{strong baseline} does not perform well since it achieves the worst performance among all the evaluated methods. The behavior of SpCL is very instructive. At the beginning of training (until the second task), the source-free model performs better but shows degraded performance later in training. This behavior can be explained by a probable divergence of the model that forgets its initial source model and overfits on the target task. On the contrary, the SpCL variant that uses the source data needs more time to handle the domain shift but keeps slowly increasing. 
Concerning the variance of the performance, we can see that the four frameworks are sensitive to their random seed, especially at the beginning of the adaptation process. However, this variance decreases after a few tasks, showing that the training becomes more stable (after two tasks for most methods) except for the \emph{Strong baseline}, where the variance of the performances becomes even higher on late tasks.

In the Market$\rightarrow$MSMT configuration (Fig~\ref{fig:MMSMT}), conclusions drastically change since SpCL
has almost the same results as MMT, hence, breaking the big gap between the two methods in performance we observed in previous configurations. This change can be explained by the large training target dataset MSMT. Therefore, every target task contains more images and more identities. This difference is beneficial to both SpCL variants that perform similarly. Regarding, MMT, we see that the performance starts degrading from task 3.  Again, it can be explained by the fact that in the case of a large target dataset, MMT can forget the knowledge from the source domain that is not further used during adaptation. Interestingly, the best performance of MMT (end of task 3) is higher than the best performance of SpCL. It illustrates the importance of handling the divergence problem and designing efficient consolidation mechanisms. Finally, we observe that the strong baseline worsens the performance compared to the initial pre-trained model.
Regarding the variance of the performances, we can see that MMT, SpCL, and SpCL-SF finally get more or less similar results at the end of the adaptation process.
Finally, in the Duke$\rightarrow$MSMT configuration (Fig.~\ref{fig:DMSMT}), the conclusions remain similar to the previous setting. Nevertheless, we can mention that SpCL outperforms MMT in this specific configuration, and observe higher instability on the SpCL-SF method that oscillates in the last tasks.

\subsection{Analyses}
\label{sec:ablation}
\begin{figure}[t]
  \centering
  \includegraphics[width=0.9\columnwidth]{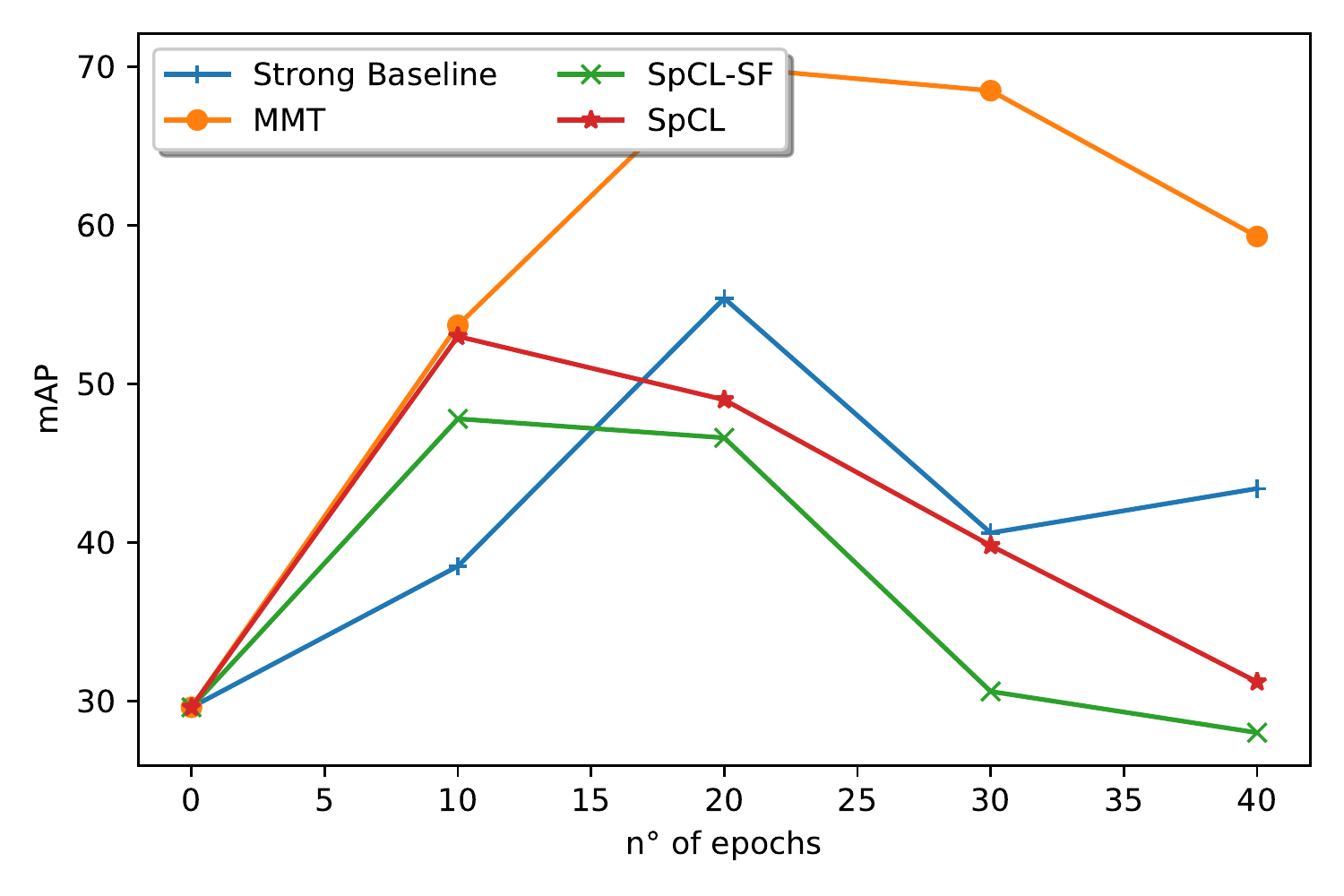}
      \vspace{-0.2cm}
    \caption{Effect of the number of training epochs per task on the Re-ID performance. At zero, we reported the results from the \emph{direct inference} model.}
  \label{fig:ablationEpoch}
\end{figure}

\noindent\textbf{Model sensitivity: number of training iterations}.
In this section, we study the effect of the number of epochs on the performance of the four frameworks (\emph{Strong baseline}, MMT, SpCL, and SpCL-SF) in the following configuration: 5-task OUDA  Duke$\rightarrow$Market.
In Fig.\ref{fig:ablationEpoch} we report the performance on the target test set (mAP) of the three frameworks while varying the number of training epochs between 0 and 40 epochs per task. Note that zero epoch corresponds to the \emph{Direct inference} performance of the pretrained model without any training on the target domain. It can be observed that with 20 epochs the \emph{strong baseline} achieves the best performance on the test set. When we increase the number of the training epochs, we see a decrease in the performance on the test set of the three frameworks probably illustrating overfitting issues in the current training task. SpCL, thanks to its memory-based system, that provides supervision from the labeled source domain images to the Re-ID model, needs fewer training epochs per task to converge, compared to the \emph{strong baseline} and MMT. We see that the four aforementioned frameworks are sensitive to the number of training epochs to some extent. These experiments illustrate the difficulty of the OUDA setting where only a few samples are available in each task and where overfitting can appear rapidly.

\begin{figure}[t]
  \centering
  \includegraphics[width=0.9\columnwidth]{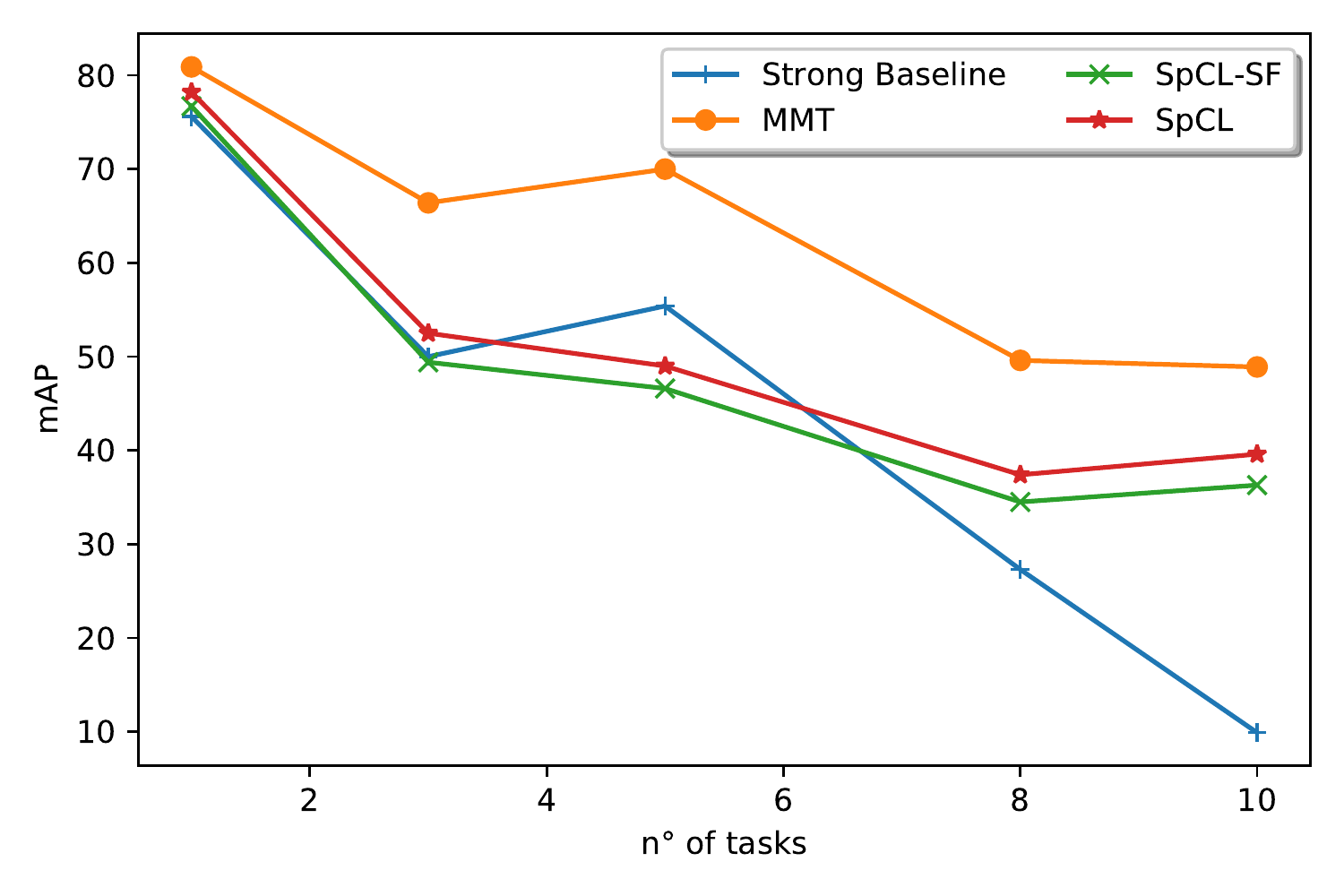}
      \vspace{-0.2cm}
    \caption{Effect of the number of tasks on the performances of the four frameworks at the end of the adaptation process. We varied the number of epochs from 1 to 10. Note that 1 epoch corresponds to the \emph{offline} setting.}
  \label{fig:multitask}
\end{figure}

\noindent\textbf{Impact of the number of tasks}. 
We also conducted further experiments to show the effects of varying the number of tasks on the adaptation performances. In Fig.\ref{fig:multitask} we report the final performance (mAP) on the target test set of the four methods when considering 1, 3, 5, 8, and 10 tasks. Naturally, when augmenting the number of tasks during the adaptation process, the number of images per task decreases. This affects the fine-tuning of the model, where we can see in Fig.\ref{fig:multitask} that for all the considered frameworks, the performance drops when considering more challenging online settings by adding more tasks.

We also performed experiments with a number of tasks larger than 10 (typically 15 or 20), however, training did not succeed due to the sampling limitation. More precisely, when the number of tasks increases the number of samples becomes too small to be handled by DBSCAN. In such challenging configurations, only a few clusters are considered, where only a few images per cluster are sampled, hence the sampling of the 16 identities with 4 images per id, which is necessary for the optimization of the triplet loss, becomes impossible. This clustering issue shows the limitation of UDA methods to address our OUDA setting and demonstrates the need for new methods tailored for OUDA.

\section{Conclusions}
In this work, we introduced the Online Domain Adaptation Re-ID problem and presented an empirical benchmark where we adapt and evaluate three state-of-the-art methods previously introduced for the \emph{Offline} UDA setting. Our experiments show that existing methods can achieve satisfactory results in simple online adaptation scenarios but fails to reach the performance achieved in the \emph{Offline} setting. We also show that the best-performing methods depend on the setting. Finally, our experiments highlight the forgetting problem when the source model is not used during adaptation. These conclusions pave the way toward novel approaches for online domain adaptive Re-ID and we hope this work will stimulate the community to address this setting that matches real-world constraints and better protect privacy.

{\small
\bibliographystyle{ieee_fullname}
\bibliography{egbib}
}

\end{document}